# ICE: Enabling Non-Experts to Build Models Interactively for Large-Scale Lopsided Problems

Patrice Simard, David Chickering, Aparna Lakshmiratan,
Denis Charles, Léon Bottou, Carlos Garcia Jurado Suarez,
David Grangier, Saleema Amershi, Johan Verwey, Jina Suh

Microsoft Research
Redmond, WA
patrice@microsoft.com

## ABSTRACT

Quick interaction between a human teacher and a learning machine presents numerous benefits and challenges when working with web-scale data. The human teacher guides the machine towards accomplishing the task of interest. The learning machine leverages big data to find examples that maximize the training value of its interaction with the teacher. When the teacher is restricted to labeling examples selected by the machine, this problem is an instance of *active learning*. When the teacher can provide additional information to the machine (e.g., suggestions on what examples or predictive features should be used) as the learning task progresses, then the problem becomes one of *interactive learning*.

To accommodate the two-way communication channel needed for efficient interactive learning, the teacher and the machine need an environment that supports an interaction language. The machine can access, process, and summarize more examples than the teacher can see in a lifetime. Based on the machine's output, the teacher can revise the definition of the task or make it more precise. Both the teacher and the machine continuously learn and benefit from the interaction.

We have built a platform to (1) produce valuable and deployable models and (2) support research on both the machine learning and user interface challenges of the interactive learning problem. The platform relies on a dedicated, low-latency, distributed, in-memory architecture that allows us to construct web-scale learning machines with quick interaction speed. The purpose of this paper is to describe this architecture and demonstrate how it supports our research efforts. Preliminary results are presented as illustrations of the architecture but are not the primary focus of the paper.

## 1. INTRODUCTION

The detection of rare concepts is helpful in several application domains. In web search, for example, we might want to distinguish book-review pages from the rest of the web. Other applications include online advertising, retail monitoring, medical imagery or telemetry. Machine Learning (ML) offers solutions to building predictive models in such lopsided problems.

Current ML solutions to these problems typically involve a "loop" of activity that works something like this. First, we collect a sample of data to label. For lopsided problems where labeling is not free, this sample is typically biased in some way to avoid having to label too many examples. To appreciate this, suppose that about one in every ten thousand web pages is a book-review page, and we sample uniformly from the web to collect data to label; if it requires on the order of a thousand examples of book-review pages to predict them accurately, then we would have to label on the order of ten million web pages before we had sufficient data. Second, after collecting the data we need to have it labeled, usually by human judges. Third, we need to design good features that the machine-learning algorithm can use to model the labels in the data. Good features for a book-review classifier, for example, might be n-gram features that are more prevalent in either book-review pages or non-book-review pages. Fourth, we train a model on one subset of the labeled data, and evaluate its performance on another subset. Depending on how satisfied we are with the performance of the model, we can loop back to the third step to choose better features (i.e., try to increase performance on the existing data), we can loop back to the first step to gather more data, possibly using our model to inform how to bias the sample, or we can decide to deploy the model in our application. After deployment, we might also decide to visit the loop again if we observe unexpected costly errors.

There are two significant problems with this methodology which limit its effectiveness. The first is that in order to be successful, a machine-learning expert is likely required in more than one of the steps: getting a biased sample to label that will result in a model that generalizes to unbiased data is a difficult ML problem, and designing features that work well might require expert understanding of the ML training algorithm. Because machine-learning experts can be hard to come by and expensive, this limits the modeling problems that are deemed worth solving.

The second problem is that iterating through the steps of the loop can take a very long time, due both to the work involved and the coordination required of the (likely many) people involved. In addition, because each iteration of the loop is costly, there is a tendency to try to minimize the number of times we iterate, which results in maximizing the number of labels we collect at each step. But labeling large batches of data at each iteration results in a loss of agility in the process. Suppose, for example, that after the first iteration through the loop, the book-review model is predicting that many magazine-review pages are being erroneously labeled as book reviews. The system designer might decide that magazine-review pages are, in fact, appropriate positives for the problem at hand and want to change the definition of the rare concept. But such a change in definition might invalidate much of the labeled data.

To overcome these obstacles, we turn to interactive ML, which interleaves human input, training, scoring, and machine feedback in a real time loop. In this setting a single teacher performs all of the functions of the domain expert, the labeler, and the ML expert. At every step, the teacher learns from the machine and the machine benefits from the human guidance. The feedback allows the teacher to gain necessary (empirical) expertise to best guide the training process without (theoretical) ML expertise. With the proper architecture, instantaneous machine training and evaluation over large datasets can shorten the typical loop above to seconds per iteration. In addition, access to large datasets enables the teacher to explore the data efficiently to surface rare examples.

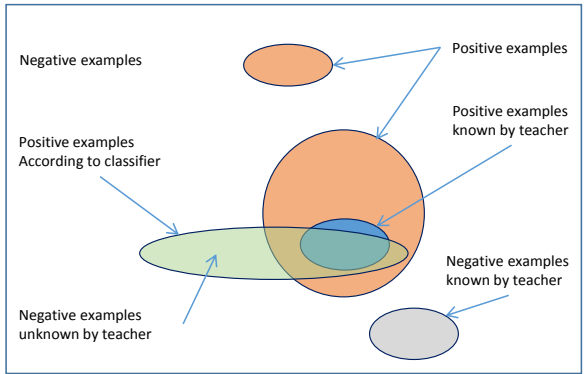

Figure 1. The world according to the teacher and the model.

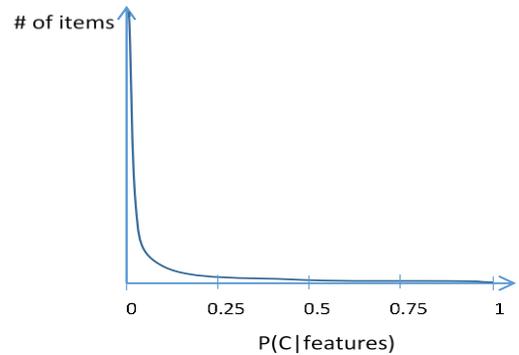

Figure 2. Typical distribution of a model score.

Figure 1 helps us visualize the "world view" of the teacher and the machine during interactive ML. Consider again the book-review classification example. Initially, from the teacher's point of view, book reviews are contained in the small blue circle and all other pages are contained in the small grey circle. We imagine a "ground truth", represented by the two orange circles, of which the teacher is not yet aware because of either ignorance or lack of clarity in the definition of a book review. From the perspective of an early stage classifier the book reviews look like the green oval. Through the process of interactive ML, the goal is to get the blue and green shapes to mold toward each other and converge on ground truth.

We built a large scale interactive ML system called ICE (Interactive Classification and Extraction) designed to leverage the capabilities of both humans and machines. Specifically, ICE enables ordinary people to train interactively both classifiers (functions that map an input item to one or more discrete classes) and extractors (functions that map an input item to a sequence of annotated segments) using large datasets containing 100 million examples or more. It allows teachers to build models on lopsided problems in a few hours without the need to pre-filter the data.

Building a large scale interactive ML system is not without its own research and engineering challenges. Interactivity requires us to train models and compute their output over large datasets in seconds. It also requires visualizations for the human to assess the model performance, understand errors, and make the right decisions in terms of the needed features or areas to explore. It further requires an interface for non-experts to label and provide features. Interactive learning also raises new questions for ML: Is it possible to derive learning algorithms whose diagnosis is easier for non-experts? Can we evaluate and even provide guarantees for an active learning scenario in which the human teacher dynamically refines the problem formulation?

The purpose of this paper is to describe the key features of our system and the corresponding architectural decisions we made to support these features. Our decisions were driven by the need to balance both (1) traditional ML objectives such as generalization accuracy, and (2) user-experience objectives such as efficiency, understandability, simplicity, and empathy with the learning algorithm.

The rest of this paper is organized as follows. First, we describe how ICE enables ordinary people to build models from scratch in a few hours. This includes describing ICE's interactive labeling and featuring loops, as well as introducing user interaction with the ICE interface. Then, we describe the distributed architecture required to enable interactive labeling and featuring at the speed and scale necessary for maintaining user engagement and effectiveness. Next, we present preliminary results on using ICE to build classifiers over web page data. Finally, we discuss open questions and related work.

## 2. ICE

In this section, we describe interactive labeling, interactive featuring and the user interface of ICE. We designed our system to allow a *single* teacher to train models interactively, and our description emphasizes this bias. As we describe below, however, we designed the system to encourage teachers to import others' models and features into their own sessions to be used as features. This functionality means that the value from a community of teachers on ICE can be much more than the sum of the value from each individual. We also envision ICE supporting cooperative teaching scenarios, where multiple teachers contribute to the same learning task [21].

### 2.1 Interactive Labeling

As discussed in Section 1, lopsided problems require us to train models on a biased sample. When a teacher first starts working with ICE (that is, when there is no model available to help choose these examples), the teacher needs to have a mechanism to reveal a-priori knowledge about the rare event in order to extract examples. In ICE, we enable the teacher to search the dataset for examples to label. Sampling by this approach alone, however, can lead to an undesirable biases due to the teacher not understanding or explicitly exploring the rare class to its full extent. For instance, the teacher might not know about or remember to retrieve all types of ethnic recipes. Consequently, a recipe classifier built only from examples found by a teacher's search might perform poorly in a real-world deployment. To deal with the teacher's ignorance about the distribution, we enable active learning to allow the system to select automatically examples that are deemed valuable to label. In this framework, teachers can be ignorant of the distribution as long as they can correctly label items on demand.

After the initial 'cold start' period when no model is available to help choose examples, ICE's interactive labeling loop alternates between the teacher labeling items and the system selecting items to label next. At any point in time, the teacher can take the initiative to search for items to label with a query. A new model is trained whenever sufficient additional labels have been submitted by the teacher. New models are immediately applied to the unlabeled data, producing a fresh score for each item. The system samples the next items to label according to the latest scores available. This active-learning strategy is efficient for labeling because it focuses the teacher's attention on items that are more likely to be mistakes and whose labels can yield better subsequent models, resulting in faster learning overall (see Section 4 on active learning). To appreciate this, consider Figure 2, which shows the number of items in the dataset as a function of the probability of a rare class according to a typical probabilistic classifier. Active-learning sampling strategies will sample "to the right" to avoid the vast majority of items that have almost no chance (according to the current model) of being a positive.

From the system's perspective, interactive labeling is a time-sensitive and demanding process. As the teacher provides labels, the system produces a model, which involves training and validating hyper-parameters (e.g. regularization parameters) in a few seconds. Automatic validation is important because non-expert teachers are not expected to provide hyper-parameters, and it is computationally intensive because it trains a family of models from which to select. Once the selected model is available, the scoring computation starts. Our infrastructure computes the score of each item, relying on parallel processes on multiple machines (see Section 3). Depending on the size of the dataset, scoring all items might not be performed in a few seconds. Even in this case, the teacher should not wait, the scores should not be outdated, and all of the data should be available to sample. We therefore keep the most recent computed score in memory for all items. We also score each item to be shown to the teacher with the latest scorer if needed. To maximize score freshness, scoring with a new model starts by interrupting the previous scorer and scoring begins where the previous scorer was stopped. Note that the quality of the selected items to be labeled next may degrade if too few items have been re-scored since the last scorer was produced. Efficient scoring in ICE is discussed further in Section 3.

Given a set of features (see Section 2.2), interactive labeling implements a fine-grained version of the traditional collect-label-feature-train-evaluate loop by updating the model within seconds after a teacher action. This means that the teacher is always confronted with the samples based on the most recent model. The teacher also has access to performance estimates for this model, which avoids the need to fix problems that have already been solved by new labels or features. Additionally, interactive labeling allows the teacher to revisit the concept definition over time without incurring such a high cost as in the traditional loop. The teacher can edit the previous labels or submit a new query to reflect this change at any point in time, which is much more flexible than (e.g.) writing new labeling guidelines to be sent along with data to a third party labeler. This highlights the benefit of interactive labeling for exploration. Our strategy not only enables the teachers to provide semantic information but also provides them with the seamless assistance of a large-scale system to evolve their mental models of the task.

## 2.2 Interactive Featuring

The performance of ML algorithms depends on the representations of the data, or features. The choice of features can greatly impact the number of training labels required to achieve a given test accuracy. The choice also affects the overall running time of training/testing and impacts the ability for human operators to understand model scores and diagnose accuracy issues. In particular, the set of features should achieve a delicate balance between (1) being representative enough so that the model can discriminate between items for the task at hand and (2) being simple enough to prevent the learning algorithm from being influenced by irrelevant distinctions between items when few training labels are available. Consequently, ML experts take great care when defining a feature set. The appropriate set is often task-specific, and its definition requires alternating between training a model, looking at its errors, and implementing or refining feature functions. This task requires both intuition about the semantics of the problem and understanding of the model errors.

ICE enables non-expert teachers to define the appropriate feature set through interactive featuring. Interactive featuring includes the creation, editing, and refinement of features. It integrates with interactive labeling in the same user interface (see Section 2.3). At any point in time, the teacher can decide to label or feature by inspecting the model performance and exploring its errors. Interactive featuring encourages the teacher to interactively compose a small set of useful features, keeping the complexity of the model to a minimum. This strategy avoids complex models with large number of features which are likely to work well in an initial phase, but may quickly leave the teacher with an intricate model that no obvious decision can improve. After providing each feature or label, the teacher can explore the model errors, interpret them and decide how to fix them through search, active learning or featuring. This decision of how to fix errors can be driven by contrasting training and test errors: training errors often can surface missing features, showing that the model is blind to some types of item characteristics, whereas testing errors can often suggest that more labels are needed for a particular area of examples.

ICE allows teachers to add three types of features: (1) built-in features, (2) features that are themselves learned *models* built by ICE teachers, and (3) teacher-defined features. Some of these features support per-item visualization to help the teacher understand their effect. Built-in features correspond to typical features used in the literature for classification and extraction, e.g. bag-of-words, regular expressions for text tasks, and edge detectors for images. Features that are learned models offer the opportunity for teachers to decompose their tasks into sub-problems. These features also offer the ability for a teacher to re-use relevant models already built by someone else. For instance, a teacher working on image classification to detect mountains can re-use or build a separate model to discriminate between indoor and outdoor pictures. Teacher-defined features includes generic functions that teachers can import by providing a dynamically linked library or features that can be edited from the ICE interface itself.

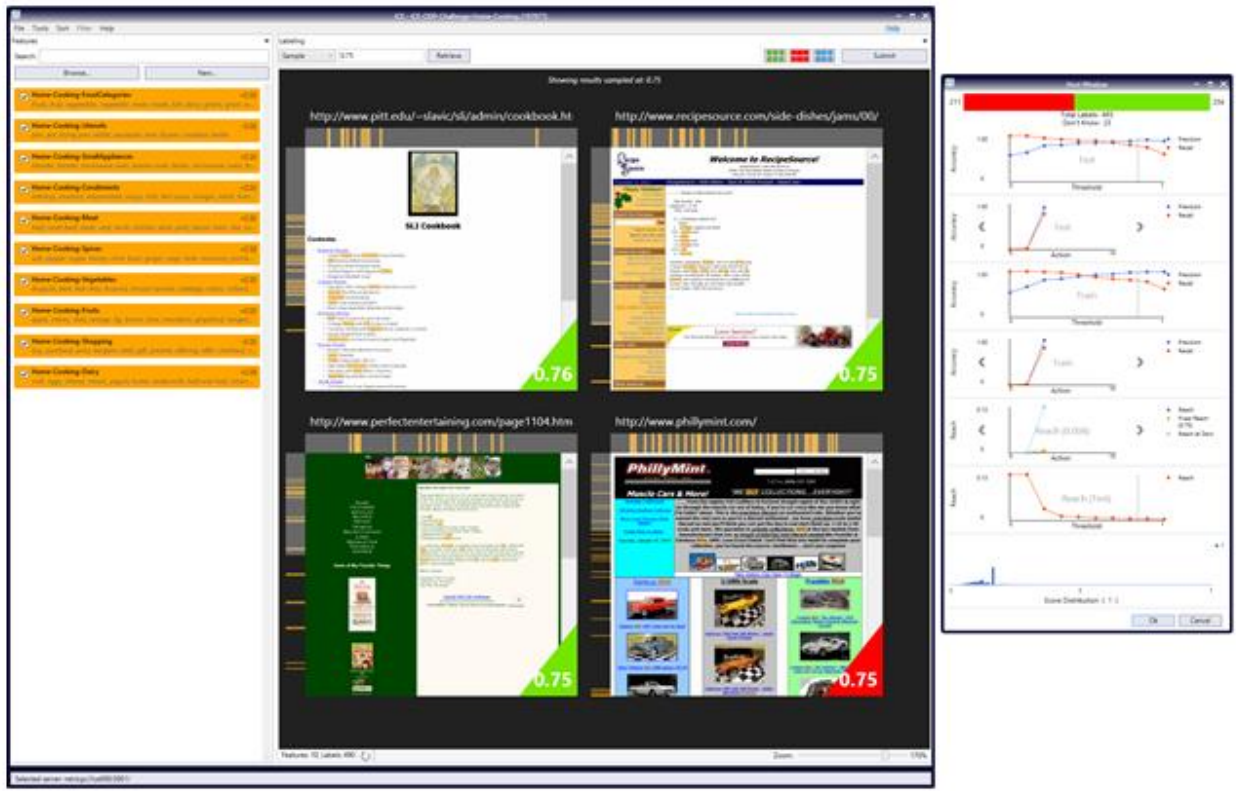

**Figure 3. The main ICE interface (left) along with a supplementary window displaying performance metrics (right). On the right of the main ICE interface, the user can inspect and label data (e.g., four pages are displayed in this figure, three labeled positive and one negative). The left panel contains features already supplied by the user for this classifier via active featuring. The supplementary window on the right display standard performance metrics (e.g., precision and recall) to help the user monitor their classifier's progress.**

In the following, we illustrate the benefit of ICE-editable features by describing *dictionaries*, a type of feature that the teacher can edit interactively for text tasks. A dictionary defines a semantic concept through a list of related words. For instance, a set of months (e.g., "January", "February",…, "December") captures the "Months" concept. The words in a dictionary together form a simple model for a token-level feature that optionally can be aggregated at a document level by computing statistics between a document and the given dictionary (e.g., how many words in the dictionary appear in the document, how many distinct words of the dictionary appear in the document). During an interactive model building session, the teacher can add, remove or edit dictionaries. We have a variety of auto-suggestion tools that help users complete partially constructed dictionaries using (e.g.) Wikipedia infoboxes. Each dictionary establishes an equivalence class over words and enables the learning of models that can generalize over words which might not even occur in training examples.

## 2.3 Interacting with ICE

ICE enables teachers with no expertise in machine learning to build classifiers and entity extractors. ICE's user interface reflects this objective and allows a few key actions that do not require ML or engineering skills. The teachers using ICE can (1) search or sample items to label, (2) label these items, (3) select and edit features, (4) monitor accuracy and (5) review errors. Training, scoring and regularizing are not teacher actions on ICE. Rather, these computations happen implicitly and transparently. Training and scoring starts automatically after modifying features or providing labels. The teachers are always aware of the state of the system as a status bar indicates which actions are not yet reflected in the current model.

Figure 3 shows the ICE interface for a classification task. When teachers start building models in ICE, they select at least one initial feature and they search for some seed positive and negative items via a text query. They can then label data resulting from the search and submit these labels (top right of main window in Figure 3). From this point, ICE automatically trains a model and starts making predictions on new data, i.e. producing scores. The teachers can then sample data that are deemed useful to improve the model (active learning), or keep searching the dataset. If a model is available (i.e., after the cold-start period), ICE pre-labels the examples shown to the teacher with the current model's most-likely prediction. As a result, the teacher can label efficiently by simply correcting those pre-labels that are not correct. Moreover, the process of explicitly correcting the model helps the teacher understand the weaknesses of the current model.

Teachers can supply features to ICE via the panel on the left in Figure 3. The teacher can either browse a corpus of existing features or create a new feature from scratch (e.g. dictionaries). Features that represent the information that the model currently 'sees' are always visible in the interface.

At any point in time, teachers can evaluate their models: ICE splits the labeled data into a training and test set so that it can compute and display performance metrics, including estimates of the generalization performance (right in Figure 3). To help the teacher monitor progress, ICE graphically displays the number of positive and negative examples provided by the teacher (to encourage a balance of positive and negatives) as well as precision and recall over time and over various prediction thresholds. Every time a new model is trained, these performance metrics are automatically updated in the interface. Teachers also have access to a review panel that shows their current labels, along with the models' predictions. The interface highlights errors and allows the teacher to sort and filter data, e.g. to view the model's predictions on the test data or to view only false positives/negatives. The review interface also enables teachers to edit their labels if they made mistakes or if their understanding of the concept of interest evolves.

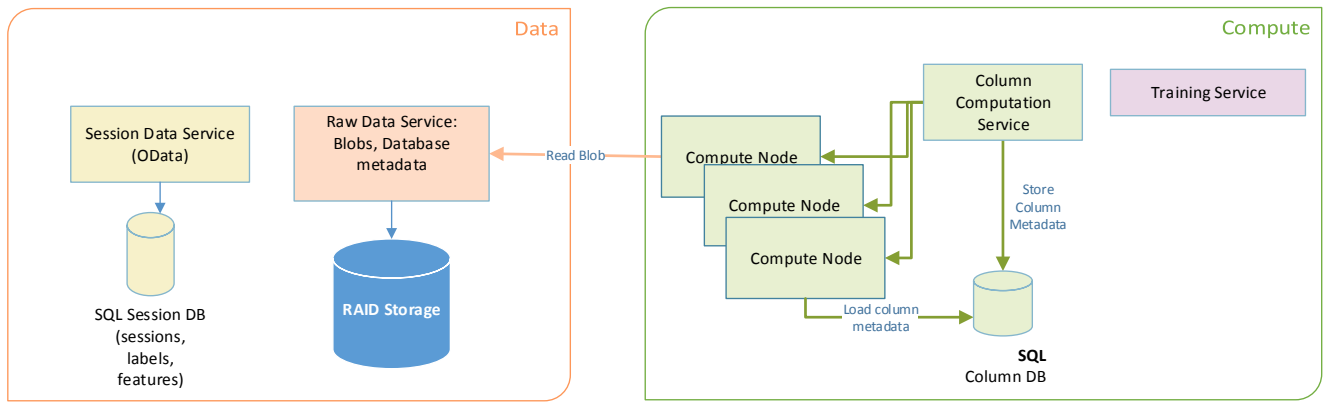

**Figure 4. ICE Services**

Teachers are therefore empowered to label, feature and search. They can understand the performance of the models they produce. When they feel confident about their model, it can be exported for deployment.

## 3. System Architecture

ICE's architecture is split into three components: (1) a teacher-facing client application for driving the ML workflows, (2) a set of services for computation and data storage, and (3) offline tools for data import.

Each component is designed to operate on arbitrary data types that need not be known in advance. In order to support this generality, the system must be extensible such that new data sets can be supported without changing the base system.

We enable extensibility by delegating some tasks to system extensions we refer to as *plugins*. Plugins extend the system on both the frontend and the on the backend. On the backend they provide a set of functions that map the raw data into features or known data types upon which features can be built (e.g. strings). On the frontend they provide visualizations that the client can display to the teacher as well as metadata about the item that can be used to enhance the visualization. For example, in the case of text, the bounding boxes of the tokens are surfaced to provide feature visualization and entity labeling.

### 3.1 ICE Client

The ICE Client is a Windows desktop application built on .NET and Windows Presentation Foundation using an MVVM (Model, View, View Model) architecture. It provides an interface for various tasks such as labeling, featuring and progress reporting. Both the user interaction and the session workflow of building models are defined and run on the client, whereas the compute or data intensive operations are delegated to backend services.

### 3.2 ICE Services

ICE provides four services that can be categorized into two types: there are two data services and two computation services (see Figure 4). The data services are realized via a machine running SQL server with a 20TB RAID 6 disk store. The two computation services are used interactively by the teacher and have stringent sub-second delay guarantees; to achieve this interactivity, the computation services are realized via a cluster of 60 machines, each with 128GB of RAM. ICE's computation services do not store any computed data permanently. Instead, they store the functions used to compute the data along with references to data stored in the data services. This allows the services to reliably regenerate any previous computation. The computation functions are submitted and stored as .NET assemblies that can be dynamically loaded. This gives teachers the ability to add functions to the system without the need to recompile or restart any processes.

In the following subsections, we provide details for each of the four services.

#### 3.2.1 Raw Data Service

At a basic level, ICE stores raw (byte array) data in a reliable way (a RAID system or cloud based redundant storage) and makes it available to the rest of the system via the Raw Data Service. The service only deals with raw data and is agnostic to higher level data types. The data is stored and organized in groups called *buckets*. Grouping the data items in buckets simplifies metadata management. In addition, it allows us to group large chunks of data for more efficient I/O.

#### 3.2.2 Session Data Service

The Session Data Service reliably stores all teacher activity associated with a model-building session; we include time stamps with every teacher event so that we can reconstruct the state of the session corresponding to any past point in time. The data is stored in a SQL database and is exposed to the client through an OData[1] API.

#### 3.2.3 Column Computation Service

The column computation service performs computations that involve the entire dataset (featuring, scoring and sampling). The service runs on a distributed column store that caches a copy of all or most of that data. Intuitively, the columns are simply the (named) values stored for each row.

---
[1] http://www.odata.org/

For the purpose of this section, *dataset* can be understood concretely as a collection of items of the same type of data (e.g., webpages, images, emails, news articles) for which features are computed. These features, in turn, are the inputs to the ML algorithm. Although (as we discuss later) we allow administrators to append data to an existing dataset, it is convenient to assume that datasets contain a fixed number of rows. The specific ordering of rows is assumed to be arbitrary but fixed so that items can be referred to uniquely by their row number (called *rowId*).

Featuring can be understood as the teacher providing a computable function of the items or other previously provided functions. In ICE, such features are special types of columns called $\lambda$-columns[2] that are maps on other columns. Scoring is also expressed as a $\lambda$-column of the features and the data that determines the models (e.g., the feature weights). Sampling (i.e., the task of selecting items to label) is expressed as an aggregation on the dataset. We also support *text queries* for certain datasets using a reverse index on the items.

We require that the column store supports random access to the rows in the dataset for two reasons. First, the training service (described in the next section) queries the column store for the feature values corresponding to the labeled items. Second, the ICE client queries the column store to extract, with low latency requirements, column values (e.g., features) for specific items that are being labeled.

As is evident from the above discussion, a general map-reduce framework (e.g., [7] or [4]) captures many of the requirements for our system. In what follows, we describe the differences between our system and existing solutions. This will help illustrate some of the architectural needs that arise from the specific use of the column store for large scale interactive ML tasks.

*In-memory column store.* An important constraint on the design of ICE is imposed by the requirement of interactivity. This means that all the actions where the teacher is sensitive to latency on the part of the system have to be fast. In particular, querying and getting items to label has to be very fast. Running aggregations corresponding to a teacher sampling the data has to be very fast. Also, scoring cannot be too slow or at least must score a diverse set of items as various iterations of the model are trained so that the teacher can inspect the output of the "current" model on a fresh set of inputs. Traditionally, map-reduce systems, such as Hadoop or Dryad ([7] , [4]) are geared towards batch processing of large data and as such make a poor fit for our interactivity needs. For ICE, we designed the column store to hold most of the columns in memory (i.e., in RAM) instead. For webpage datasets, for example, we hold the text content (and layout geometry information) of all webpages in memory on the cluster. Furthermore, feature columns and scores are also held in RAM. There are certain columns, such as the rendered images of webpages that are too large to hold in RAM for the size of datasets we consider. The teacher typically inspects only a few of these in the course of labelling. Thus we load these on-demand from the disk. More recent in-memory map-reduce systems such as Spark [23] are more similar to the ICE column store. The differences with such systems have more to do with trade-offs we make in ICE so that certain operations that are needed for interactive learning are even faster at the cost of less generality.

---

[2] The terminology is from lambda calculus.

*Data Layout.* A column is split into multiple chunks, called *buckets*, of contiguous rows; these buckets are distributed across many machines. The data layout has the additional property that all the columns corresponding to the same row are present on the same machine. This design decision imposes the constraint that all the information needed for a model to score a single item is present on the same machine. In practice this is not a limitation because the feature vectors are not too large. Each bucket stores the starting *rowId* of the items in the buckets and one can compute the *rowId* for any item in the bucket by examining the offset of the item. Metadata specific to a dataset gives the mapping of intervals of *rowId*s to machines.

*Aggregations.* ICE supports global associative aggregations and joins on the `rowId` key and more general reduce operations[3] are not supported. This is a simplification that arises as a consequence of ML tasks generally either needing to sample the data or compute statistics (e.g., TF-IDF) over the dataset. Such aggregations can also be run on multiple columns simultaneously. An example is an aggregation where an item is picked uniformly at random from unlabeled data with a score between 0.70 and 0.75 and a particular feature is also present. In this case we would aggregate the score column and feature column, ignoring rows where the feature is absent, and do a reservoir sampling for items with a score in the desired range. ICE allows for the results of an aggregation to be wrapped in a column. This means that the aggregation result is repeated for each row in the column (though it is not stored this way to conserve space). This is especially useful when the results of an aggregation are used in a $\lambda$-column.

*Reverse index for text searches.* Labelling for text classification tasks benefit from supporting literal text queries. This allows for efficiently finding examples in the positive class. In ICE we enable this by allowing a reverse index on certain string columns. These indices are built at a machine level. For text queries the results from each of these machines are ranked based on their Okapi BM25 score (see [15] §11.4.3).

*Scoring.* In ICE the score column is special and we explain how it is treated differently first. As the teacher labels new data, the classifiers or extractors are retrained and the items are re-scored with the latest model. The new scores are used to select which item to label next (e.g., items whose probability or being positive is between 0.7 and 0.8).

One problem that arises when successive models are built in a session is that scores from one model may not be comparable to scores from another. Having comparable scores can be important for sampling correctly, and so we include support for automatically calibrating the scores. For example, when the score represents a probability of an item belonging to a class, we use isotonic regression to keep these scores calibrated.

Another problem that arises is computational. Scoring is often much slower than labeling or re-training because it can involve scoring the full data set. This means that the teacher would have to wait for scoring to complete if only the scores from the latest scorer were to be used. To circumvent this problem, the score columns can contain values that have been scored by different versions of the model. Intuitively, these columns can be thought of as having their values arranged in a circular list, and the latest available model is used at all times to overwrite these values.

---

[3] The limitation to global associative reduce operations (i.e., associative aggregations) is similar to the original design decision in Spark [23]; more recent implementations of Spark allows general reduce operations (and even DAG workflows [24]).

Aggregators have access to the version number of the model used to produce the (latest) score of each item, which allows much flexibility for sampling algorithms during the interactive loop.

Despite these latency-hiding techniques, the speed of scoring remains important because scores from more recent models give better feedback to the teacher. Having more machines results in fresher scores and more efficient labels, but the behavior degrades gracefully when fewer resources are available. The mechanism that ICE uses to increase the number of items that have fresh scores is a distinguishing feature of ICE in relation to general purpose systems like Spark.

### 3.2.3.1 Fault tolerance

As in any distributed system, failures are expected. ICE's approach to fault tolerance is specifically designed to support its data availability requirements. The Column Computation Service allows access to data in two distinct ways: a) fetching specific row values for a column and b) aggregate operations over the entire data set, such as sampling queries, aggregation computations (e.g. IDFs) or text queries. Having specific row values always available is important because they may be needed to present an item to the teacher, or to collect feature values for training. In contrast, aggregate operations can be robust to a portion of the data missing. For example, sampling by score or searching can yield good results even if they are not run over the entire data set.

For this reason, ICE only guarantees availability of all the data for single row accesses. If a compute node goes offline and a row value is needed from data hosted by this machine, the request is redirected to another random compute node. This alternate node fetches the raw data for this row from the Raw Data Service and any lambda computations needed to produce the value are executed on demand. While this is much slower than accessing pre-computed values cached in memory, the number of rows for which this needs to happen is generally small, which makes this a viable solution.

### 3.2.3.2 Benchmarks

ICE has been tested with a data set that consists of 35 million web pages deployed across 35 compute nodes. The pages are saved in a proprietary format that includes a snapshot of the rendered page along with all the text and DOM structure information. This data set occupies 4.5 TB of disk space and a separate version without the page images is also maintained for faster access and occupies 710 GB. The extracted text and html geometry measures around 182 GB and this is the portion that is kept in memory on the compute nodes.

In terms of computation speed, running a linear classifier with dictionary features (pre-computed) takes approximately 2 seconds. This corresponds to scoring >40 000 documents per second per core in the cluster and is an area of the system than can be further optimized.

As mentioned earlier, the most important aspect is querying speed. The system can scan classifier scores for all 35 million pages in under 250 ms, which gives good interactive responsiveness to the teacher.

For comparison purposes, we ran a map/reduce test in both ICE and Spark[4]. The task consisted of a map counting the number of occurrences of a word in 400 000 pages from the Open Directory Project[5], followed by a sum global reduction. This was done on an 8 core system with 14 GB of RAM (the data for this test is 1.2 GB). The following table summarizes the results and shows equivalent performance across both systems (times in milliseconds):

|  | Spark | ICE |
|---|---|---|
| Map (count) | 3840 | 3500 |
| Reduce (sum) | 55 | 5 |

### 3.2.4 Training Service

The Training Service is responsible for hosting a catalog of *trainers* (e.g., model-building algorithms). These trainers can be grouped into *families*, which correspond to groups of instances of these algorithms configured with different hyper-parameters. An ICE session uses the training service by selecting a *trainer* or *family* and requesting that it is run on its particular training set. In the case of a family, the training service can run all member algorithms in the family and select the best model based on cross-validation of a given metric (e.g. AUC).

## 3.3 Data Import Tools

Bringing data into ICE requires an administrator to provide an importer object (in a dll) that can enumerate the individual rows of some raw-data source. Importantly, the details/types of this data do not need to be known by ICE; rather, the corresponding labeling and featuring *plugins* are the only components that need to be able to translate the raw data into types (labels and features) that are understood by ICE. An important aspect of ICE data sets is that they are *append only*. New items can be added, but previously imported items are kept in immutable form. Immutability of the raw data is key to being able to guarantee immutability of the columns that are computed from it.

## 4. Results

In this section, we present some preliminary results from people using ICE to build classifiers on standard text classification tasks.

## 4.1 Experiments

Recall that our goal with ICE is to enable everyday people to train ML models from scratch over big data. This requires balancing traditional ML performance metrics (such as generalization accuracy) with user-experience metrics (such as efficiency, understandability, empathy with the learning algorithm, and simplicity). Therefore, here we present results pertaining to both. To create our ground truth, we gathered webpages with labels provided by the Open Directory Project[5] (ODP) and crawled the URLs to get the HTML and text. Any pages that timed out, were redirected, or which contained fewer than 100 characters were excluded from the set. The resulting truth set was split randomly into a 70% train set with 330,398 examples and a 30% test set with 140,839 test examples.

---

[4] We chose Spark for this comparison because that is the system closest to ICE in capabilities.

[5] http://www.dmoz.org/

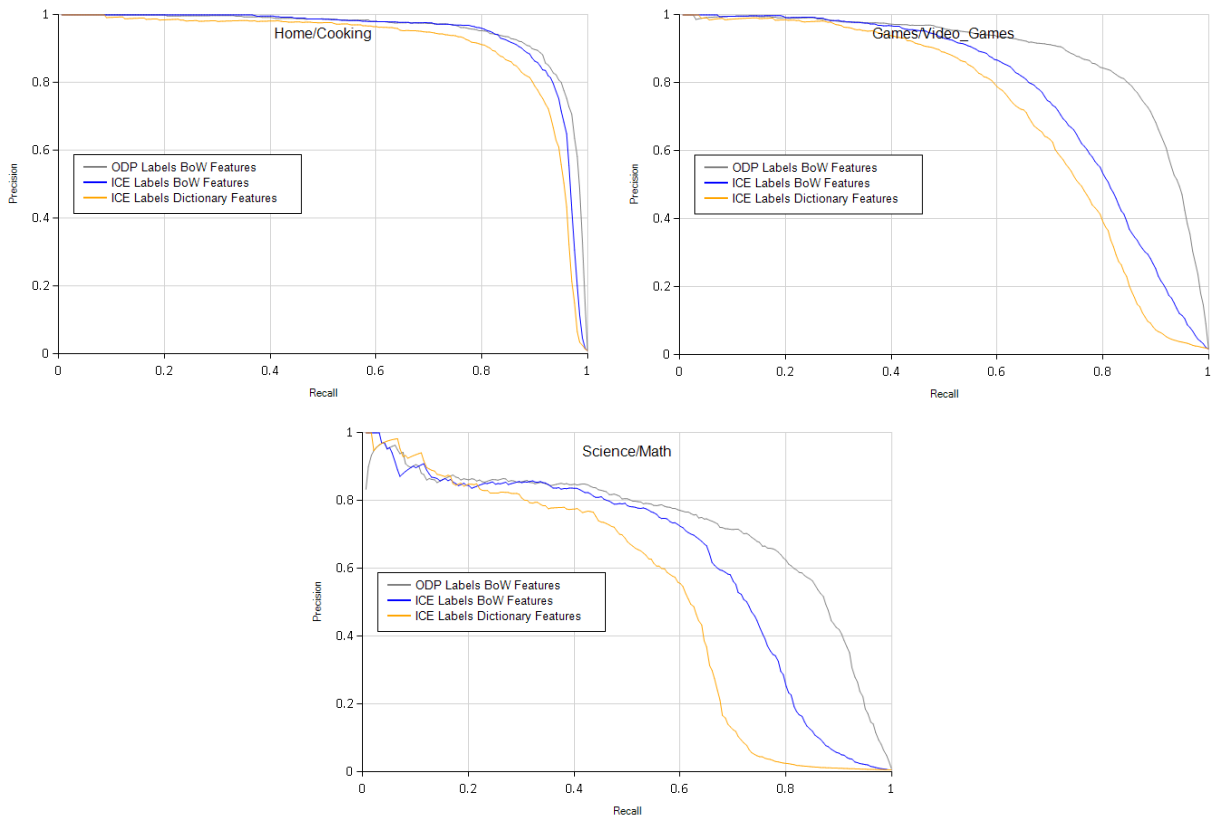

**Figure 5. Performance of ICE built classifiers compared to classifiers built via traditional means (i.e., a large labeled data set using bag-of-words features).**

Our participants for this experiment were members of our team who selected concepts from the ODP hierarchy based on their familiarity with the concept. While some members of the team qualify as ML experts, that knowledge did not give them any specific advantages. Indeed, the best classifiers were not built by ML experts. The concepts selected were: Home/Cooking, Science/Math, and Games/Video Games. The ground-truth number of positives for each of these categories ranged from 1% to 8% of the entire corpus and were equally represented in the train and test sets (8% for Home/Cooking, 6.5% for Science/Math and 1.7% for Games/Video Games).

We required participants to start from scratch and use ICE to interactively build a web-page classifier for their selected concepts via ICE's interactive ML process described in Section 2. Teachers were not shown the true ODP labels for this exercise. Teachers created dictionary features based on what they deemed important for predicting their categories. For example, one participant created dictionaries of "ingredients", "utensils", and "measurements" for the Home/Cooking classifier.

When participants were done, we analyzed their sessions to compute the total time taken, the number of labels and features provided, and the resulting classifier performance. We also computed these metrics for a classifier trained using the same labels, but after replacing the teacher-supplied dictionaries with a standard bag-of-words (BOW) feature set consisting of the top 10,000 n-grams from the Google Web n-grams and weighted by TF-IDF. We compare these "ICE labels" classifiers to a baseline "ODP labels" classifier trained using all of the ODP ground truth labels. In all cases, the classifiers were trained with logistic regression using an L2 regularizer.

Figure 5 show the results from this experiment. We see that within a few hours of work (3-8 hours) and a few hundred labels, teachers are able to create classifiers using ICE (orange and blue curves for dictionary and BOW features, respectively) that match traditionally built classifiers (gray curve) especially in the high precision region of the precision/recall (P/R) curve. Table 1, summarizes these results, showing that participants were able to create classifiers in a few hours with a very small number of labels (on the order of 100s) and a small number of dictionary features (30-70 compared to ~300,000 labels and 10,000 features using the standard BOW model). In addition, reducing the number of labels and using teacher supplied conceptual dictionaries both contribute to the simplicity and interpretability of the models. Whereas ICE classifiers built in a few hours give teachers a quick way to get several weak classifiers, the result also shows that the ICE labels with BOW features outperforms ICE classifiers built with ICE labels and dictionary features. In some of these cases, the dictionaries built by teachers were either noisy or incompletely capture the intended concept (e.g., the classifier built for Video Games missed a number of positive pages talking about "clans" since the teacher did not know that "clan" referred to a group of people playing Video Games). In the next section we discuss some of these issues and briefly describe our ongoing research on improving the performance of ICE features.

| | Home/Cooking | Science/Math | Games/Video |
|---|---|---|---|

|  | ICE | ODP | ICE | ODP | Games ICE | ODP |
|---|---|---|---|---|---|---|
| Time (hours) | 6.87 | NA | 5.42 | NA | 2.64 | NA |
| Num. Labels | 767 | 330,398 | 588 | 330,398 | 406 | 330,398 |
| Num. Features | 68 | 10,000 | 66 | 10,000 | 32 | 10,000 |
| Recall at 80% Precision | 90% | 93% | 31% | 50% | 43% | 82% |

**Table 1: Table summarizing the labels and features used in ICE classifiers vs. a standard BOW model**

## 4.2 Deployment

The ICE tool is being used by several customers from Microsoft business units with the functionality mentioned above (interactive labeling, featuring, sampling, and reviewing) to build classifiers and entity extractors. Several of the customers are non machine learning experts and the classifiers they have created with ICE have been deployed in commercial applications. An example of this is the use of ICE in the live Cortana service. Here we apply ICE to build classifiers for the social conversation domain where a domain expert in the Cortana team (non machine learning expert) developed a high precision binary classifier in 135 minutes to detect the COMPLIMENT intent, for utterances such as "Cortana, you are great". Social conversation is one of several customers of a system like ICE. Other applications include web search, email classification and telemetry.

## 5. Related Work

ICE builds upon recent advances in interactive machine learning and distributed systems. The following presents work relevant to ICE.

## 5.1 Interactive Machine Learning

Different fields have independently introduced systems that iterate through labeling, model refinement and output updates in an interactive way. Such systems can target non-technical users. For example, relevance feedback mechanisms in Information Retrieval is one of the earliest form of interactive learning [17][18]: in a retrieval scenario, the user receives a document ranking from an initial query and marks some of the documents in that ranking as relevant or non-relevant. The system then takes this feedback into account to compute a new ranking. Similar mechanisms have also been proposed for recommender systems [1][16]. In these cases, the end goal is not to produce a model but rather to assist quickly the user to find a small set of relevant items.

The idea of interactive model building grew with the field of active learning [19]. Active learning studies learning algorithms and sampling strategies to learn models by actively querying human labelers. It aims at understanding the impact of biased sampling on generalization performance and proposing practical solutions with theoretical convergence guarantees. Theoretical works percolated to practical applications. Interactive training has been applied to computer vision tasks such as image segmentation [8], image classification [10], or gesture classification [9]. Interactive learning has also been applied to text classification tasks [11]. In particular, different forms of interactive featuring have been explored over text data [2][6][20].

ICE differs from prior work by its generality as it can be applied to different data types (e.g. web-pages, raw text or images) and ML tasks (e.g. classification or segmentation). It also differs by the dataset size it can manipulate. To our knowledge, ICE is the only active learning system that enables building models over 100 million items with interactive speed.

## 5.2 Distributed Systems for ML

Distributed systems for machine learning commonly refers to parallel learning algorithms designed to carry the training optimization over large datasets, e.g. map-reduce for expectation maximization [5] or parallel asynchronous stochastic gradient [14]. Our application is different. Training is typically inexpensive and performed on a single machine because ICE's training sets are limited to the labels produced by one user over an interactive session. Our challenge resides in model scoring. The interactive loop frequently produces new models, and we require the system to compute quickly model outputs over large datasets for active sampling. Our computational requirement is therefore similar to the one driving the design of Spark [23], a cluster computing system for interactive data analysis as discussed in Section 3.

## 6. Discussion and Future Work

High performing classifiers and extractors on lopsided problems are important. The traditional approach to create them is inefficient and requires hard-to-come-by (and expensive) ML expertise. To improve on cost, speed, and expertise requirements, we turned to interactive learning and built an environment that enables *teachers* to quickly interact with data at scale with ML in the loop.

The resulting platform, ICE, is being used for two purposes: to allows ordinary people to build ML models over big data from scratch and in a few hours, and to enable researchers to explore the challenges of interactive learning. The paper presents the important choices we made to create an interactive ML architecture, and some results we obtained from non-expert early adopters.

The architecture of ICE is very similar to the architecture of Spark. This is remarkable because ICE was designed specifically for interactive learning and Spark was designed with a more general ML platform in mind. The two designs share a large distributed immutable in-memory column store. Spark implements an in-memory version of Map-Reduce. ICE supports a limited version of Map-Reduce, without GroupBy, which permits every column of a given row to reside on the same machine and reduces the intra-cluster communication. ICE uses this property to implement a highly optimized instance of Map-Reduce dedicated to filtering by score (for active labeling and exploration). The ICE map-reduce is available to engineers building the datasets, but not to the teachers. A low priority Map is available to the teacher for computing new features on the large dataset (computing features on labeled sets remains very fast). These policies are in place to ensure that all available computation power is used to minimize the response time from the teacher's point of view. Despite these differences, it is possible that Spark or a similar platform could be used by ICE in future iterations.

Classifiers built in a few hours by non-experts using the ICE perform reasonably well compared to the state of the art. Our ultimate goal, however, is to enable anyone to build models in a very short time that perform as well or outperform models built by ML experts via traditional means.

Once the paradigm of interactive ML is adopted, we can think of many directions of research to improve the learner, the teacher, and/or their interaction. For instance, a promising approach that does all three is to enrich the feature *language* used by the teacher to tell the model what to observe. Preliminary results indicate that we can extend the user-supplied dictionary features (for text domains) beyond literal string match to create significantly more powerful and informative features. Consider, for example, a dictionary feature that contains the abbreviations for the US states; whereas the user had the concept "US state" in mind, a literal string match will result in the feature being on for every instance of "in" (Indiana) and "or" (Oregon), which is likely to diminish the utility of the feature. As an alternative, we are using the entries in each dictionary to build automatically (i.e., from unlabeled data) a common *context model* that identifies surrounding string contexts where multiple dictionary entries have high probability of appearing; by shifting from literal strings to context models, the dictionary features can match strings not explicitly given (e.g., common misspellings) and can non-match on strings explicitly given (e.g., when "in" is not used as a state abbreviation). On the few classification examples we investigated, these "concept" dictionaries have proven to be far superior to "literal" dictionaries because they only need few words to define a common context (efficient), they are not blind when none of the words in the dictionary appear in the document (better recall), and they seldom fire in the wrong context (better precision).

Another promising research direction is to provide tools to enhance the teacher's ability to do exploration. For instance, different metrics (in exotic feature space) or external information sources like a click graph from a search engine (connecting query nodes to document nodes) can be used to suggest positives for queries or for web pages.

## 7. REFERENCES

[1] Adomavicius, G. and Tuzhilin, A. 2005. Towards the Next Generation of Recommender Systems: A Survey of the State-of-the-Art and Possible Extensions. *IEEE Transactions on Knowledge and Data Engineering 17*, 6, 734-749.

[2] Attenberg, J., Melville, P., and Provost, F. 2010. A Unified Approach to Active Dual Supervision for Labeling Features and Examples. 2010. *Proceedings of the European Conference on Machine Learning (ECML-10).*

[3] Cesa-Bianchi, N. and Lugosi, G. 2006. *Prediction, learning, and games*. Cambridge University Press.

[4] Chaiken, R., Jenkins, B., Larson, P.-A., Ramsey, B., Shakib, D., Weaver, S. and Zhou, J., SCOPE: Easy and Efficient Parallel Processing of Massive Data Sets, *Proceedings of VLDB, 1(2), 2008, 1265-1276.*

[5] Cheng-Tao Chu, Sang Kyun Kim, Yi-An Lin, YuanYuan Yu, Gary Bradski, Andrew Y. Ng, Kunle Olukotun Map-Reduce for Machine Learning on Multicore. 2006. *In Proceedings of Neural Information Processing Systems (NIPS)*

[6] Druck G. and A. McCallum A. 2011. Toward interactive training and evaluation, *Proceedings of the ACM Conference on Information Knowledge Managment (CIKM'11).*

[7] Dean, J. and Ghemawat, S. 2004. MapReduce: Simplified Data Processing on Large Clusters. *Sixth Symposium on Operating System Design and Implementation (OSDI'04)*, 137-150.

[8] Fails, J.A. and Olsen Jr., D.R. 2003. Interactive Machine Learning. In *Proceedings of the International Conference on Intelligent User Interfaces.* IUI 2003. ACM, 39-45.

[9] Fiebrink, R., Cook, P.R., and Trueman, D. 2011. Human Model Evaluation in Interactive Supervised Learning. In *Proceedings of the SIGCHI Conference on Human Factors in Computing Systems*. CHI 2011. ACM, 147-156.

[10] Fogarty, J., Tan, D., Kapoor, A., and Windor, S. 2008. CueFlik: Interactive Concept Learning in Image Search. In *Proceedings of the SIGCHI Conference on Human Factors in Computing Systems*. CHI 2008. ACM, 29-38.

[11] Godbole, S., Harpale, A., Sarawagi S., and Chakrabarti, S. 2004. Document classification through interactive supervision of document and term labels. I*n Proceedings of ECML/PKDD '04.*

[12] Kapoor, A., Lee, B., Tan, D., and Horvitz, E. 2010. Interactive Optimization for Steering Machine Classification. In *Proceedings of the SIGCHI Conference on Human Factors in Computing Systems.* CHI 2010. ACM, 1343-1352.

[13] Lai, T. L. and Robbins, H. 1985. Asymptotically Efficient Adaptive Allocation Rules. *Advances in Applied Mathematics*, 6:4-22.

[14] Langford, J., Smola, A.J. and Zinkevich, 2009. M. Slow learners are fast. *In Proceedings of Neural Information Processing Systems (NIPS)*

[15] Manning, C., Raghavan P., Schutze, H., *an Introduction to Information Retrieval,* Cambridge University Press, 2009.

[16] Montaner, M., Lopez, B., de la Rosa, J.L. 2003. A Taxonomy for Recommender Agents on the Internet. *Artificial Intelligence Review 19*, 285-330.

[17] Rocchio, J. 1971. Relevance Feedback Information Retrieval. In *The SMART Retrieval System: Experiments in Automatic Document Processing,* G. Salton, Ed. Prentice-Hall, Englewood Cliffs, NJ, 313-323.

[18] Salton, G. and Buckley, C. 1990. Improving Retrieval Performance by Relevance Feedback. *Journal of the American Society of Information Science 41*, 288-297.

[19] Settles, B. 2012. *Active Learning*. Morgan & Claypool

[20] Settles, B. 2011. Closing the Loop: Fast, Interactive Semi-Supervised Annotation with Queries on Features and Instances. *In Proceedings of Empirical Methods in Natural Language Processing (EMNLP)*

[21] Azari Soufiani, H., Chickering, D., Charles, D., and Parkes, D. 2014. Approximating the Shapely Value via Multi-Issue Decompositions. *In Proceedings of the 13th International Conference on Autonomous Agents and Multiagent Systems (AAMAS 2014).*

[22] Talbot, J., Lee, B., Kapoor, A., and Tan, D.S. 2009. Ensemble Matrix: Interactive Visualization to Support Machine Learning with Multiple Classifiers. In *Proceedings of the SIGCHI Conference on Human Factors in Computing Systems.* CHI 2009. ACM, 1283-1292.

[23] Zaharia, M., Chowdhury, M., Franklin, M., Shenker, S. and Stoica, I., Spark: Cluster Computing with Working Sets, 2nd USENIX Workshop on Hot Topics in Cloud Computing (HotCloud'10), 2010.

[24] Zaharia, M., Chowdhury, M., Das, T., Dave, A., Ma, J., McCauley, M., Franklin, M., Shenker, S., Stoica, I., Resilient distributed datasets: a fault-tolerant abstraction for in-memory cluster computing, *9th USENIX Symposium on Networked Systems Design and Implementation*, 2012.